**RESEARCH PAPER**

# FL-GUARD: A Holistic Framework for Run-Time Detection and Recovery of Negative Federated Learning

Hong Lin[1] · Lidan Shou[1] · Ke Chen[1] · Gang Chen[1] · Sai Wu[1]

## Abstract

Federated learning (FL) is a promising approach for learning a model from data distributed on massive clients without exposing data privacy. It works effectively in the ideal federation where clients share homogeneous data distribution and learning behavior. However, FL may fail to function appropriately when the federation is not ideal, amid an unhealthy state called *Negative Federated Learning* (NFL), in which most clients gain no benefit from participating in FL. Many studies have tried to address NFL. However, their solutions either (1) predetermine to prevent NFL in the entire learning life-cycle or (2) tackle NFL in the aftermath of numerous learning rounds. Thus, they either (1) indiscriminately incur extra costs even if FL can perform well without such costs or (2) waste numerous learning rounds. Additionally, none of the previous work takes into account the clients who may be unwilling/unable to follow the proposed NFL solutions when using those solutions to upgrade an FL system in use. This paper introduces FL-GUARD, a holistic framework that can be employed on *any* FL system for tackling NFL in a run-time paradigm. That is, to dynamically *detect* NFL at the early stage (tens of rounds) of learning and then to activate *recovery* measures when necessary. Specifically, we devise a cost-effective NFL detection mechanism, which relies on an estimation of performance gain on clients. Only when NFL is detected, we activate the NFL recovery process, in which each client learns in parallel an adapted model when training the global model. Extensive experiment results confirm the effectiveness of FL-GUARD in detecting NFL and recovering from NFL to a healthy learning state. We also show that FL-GUARD is compatible with previous NFL solutions and robust against clients unwilling/unable to take any recovery measures.

## 1 Introduction

Federated learning is a distributed learning paradigm in which multiple devices (also called *clients*) learn a shared global model collectively without disclosing their private

✉ Lidan Shou
  should@zju.edu.cn

  Hong Lin
  honglin@zju.edu.cn

  Ke Chen
  chenk@zju.edu.cn

  Gang Chen
  cg@zju.edu.cn

  Sai Wu
  wusai@zju.edu.cn

[1] The State Key Laboratory of Blockchain and Data Security, Zhejiang University, Hangzhou 310027, Zhejiang, China

data [1]. In the vanilla FL (FedAvg [1]) system, the global model is leaned by iterative rounds, with each round containing two steps: (1) Training. The clients train the global model on their local data and then submit their local updates to a central server. (2) Aggregation. The server aggregates local updates and distributes the aggregated model as the new global model to all clients. Such a distributed learning paradigm has found a wide range of applications where data are decentralized and data privacy is essential [2, 3].

Despite the advantage of FL in protecting privacy, massive experiments on real-world datasets [4–8] have reported the *failure* of FL [9, 10]. That is, for most clients, the model produced by FL, when tested on a client's private data, cannot achieve a better performance than the *private model* produced by that client's local stand-alone training.[1] Issues leading to FL failure include (but are not limited to):

- data heterogeneity among clients, also called non-IID data [11];
- client inactivity due to connection problems or hardware failure [12];
- attacks from the malicious clients/server, e.g., data poisoning [13] and model poisoning [14, 15];
- noises introduced by privacy-protection measures, e.g., differential privacy [16].

Consequences of FL failure include: (1) clients being unwilling to participate in FL, (2) wasted rounds of client computation (and client–server interactions), and (3) disintegration of the entire federation in the worst case.

Many remedy solutions have been proposed to prevent FL failure [9, 12, 17–27]. However, the FL system using the existing solutions faces a *dilemma*: If the remedy is predetermined to be used, it incurs extra (high) costs even if FL could have done well without such a remedy. In contrast, if the system chooses not to activate the remedy first, then the possible failure and the need for the remedy would manifest only after hundreds of learning rounds. Thus, the waste of client computation cannot be avoided. None of the two choices is perfect unless *the necessity of a remedy is known in run-time*. However, no previous work addresses this dilemma. Additionally, when employing a remedy solution to upgrade an FL system in use, none of the existing work considers the following realistic scenarios: (1) Some clients are unwilling to take the proposed remedy solutions until observing the others benefit from doing so, and (2) Some clients are unable to take any remedy solution due to their limited computing/communicating power.

Many questions still remain open in the search for the cure of failed FL. For example: How to define a failed FL process? Is failure detectable at an early stage (so that numerous futile learning rounds can be saved)? Is it possible to achieve good local performance on clients when the global FL keeps failing? If yes, what are the countermeasures to be taken by clients? What are the expenses of those countermeasures? What happens if not all clients take such measures?

In this paper, we attempt to answer the above questions. With a trivial assumption that each client of the federation has a private model previously trained on its local data, our work aims at learning a model for each client that can outperform its private model. We coin a new term *Negative Federated Learning (NFL)* to name the undesirable state of an FL system in which the iterative client–server interactions do not help most clients in learning. Based on the vanilla FL paradigm, which is widely used in practice, we propose a novel FL framework called FL-GUARD for tackling NFL in a run-time detection and recovery paradigm.

Our run-time NFL detection scheme relies on a metric called *performance gain* (PG), i.e., the improvement in learning accuracy that the clients obtain from participating in FL. Measuring the PG on a client needs the accuracy of the model learned in the FL system (e.g., the global model learned in the vanilla FL system) on that client's testing dataset. However, testing a model in each round of FL may incur non-trivial extra costs for clients. To avoid such costs, we choose to *estimate* the PG instead. Specifically, each client leverages its training data as a *surrogate* to estimate the accuracy of the model learned in the FL system on its testing dataset. Then, the estimated PG is obtained by calculating the difference between (1) the estimated accuracy of the model learned in the FL system and (2) the accuracy of the private model obtained before FL starts. The estimated PG is uploaded to the server together with the respective local updates. Next, the server computes an overall performance gain during aggregation. If the overall PG value remains negative after certain rounds of learning, the system reports the state of NFL. Our detection scheme is cost-effective since the per-client PG is a numeric value that can be transmitted at a trivial cost and easily estimated when clients train the model in FL.

Once NFL is detected, the system has to take recovery measures to improve the performance of FL on each client. Our key idea is to personalize the global model learned by vanilla FL. Specifically, each client learns in parallel an *adapted model* when training the global model. We expect the adapted model can fit the local data distribution and gain benefits (e.g., generalization ability) from the global model learned in each round of FL. Therefore, the optimization goal of the adapted model is to minimize two terms, i.e., *the loss on the local data* and *the parametric divergence from the global model*. However, when the global model cannot bring any valuable knowledge for learning the adapted model, adding the second term may negatively influence model adaptation. To mitigate such negative impacts, we introduce a parameter $\lambda$, which is tuned dynamically during training, to control the weight of the parametric divergence. Tuning $\lambda$ dynamically in run-time helps to avoid the costs for the repeated tuning of hyperparameters before the start of each FL task, which is a main disadvantage of many existing adaptation methods [10, 12, 19–21, 28].

The contributions of our paper are summarized as follows:

(1) We propose FL-GUARD, a holistic framework for tackling NFL in a *run-time detection and recovery* paradigm. The framework can be easily employed on any FL system to detect NFL and recover from it. To the best of our knowledge, this is the first dynamic solution for tackling NFL in run-time. The importance of a run-time NFL solution is twofold. On the one hand, it does not incur extra (high) costs when vanilla FL performs well. On the other hand, it can save numerous training rounds conducted in vain when FL indeed needs recovery.

---

[1] Local stand-alone training means that a client independently learns a private model on its training dataset without cooperation from the others.

(2) We design a cost-effective mechanism for run-time NFL detection. The proposed scheme can detect NFL at a *very early stage of learning* and thus save hundreds of futile learning rounds before a necessary recovery is activated.

(3) We also introduce an NFL recovery method, which, activated in run-time as per result of NFL detection, improves the performance of federated learning on individual clients by learning an *adapted model* for each client.

(4) We conduct extensive experiments on federated image classification and language modeling tasks. The results confirm the effectiveness of FL-GUARD in detecting NFL at the early stage of learning and recovering from NFL. We also demonstrate that FL-GUARD is compatible with previous NFL solutions and robust against clients unwilling/unable to take any recovery measures.

## 2 Related Work

Federated learning, originally developed by Google, is an emerging technique for learning from decentralized data [1]. A significant incentive for a client to engage in FL is to obtain a model better than the private model that it can train independently without cooperation from other clients [10, 19]. However, many studies report that such an incentive is not always guaranteed. Many issues can pose *negative effects* on FL. Zhao et al. [29] observed the accuracy of a model learned by FL may drop over 50% when data distributions differ across clients. Both McMahan et al. [1] and Briggs et al. [30] showed that client inactivity would harm the convergence of federated learning. Bhagoji et al. [14] presented that attackers could easily manipulate the model learned by FL to generate false predictions. Yu et al. [10] argued that the differential privacy could also induce a significant performance drop in the model learned by FL. Many real-world FL tasks reportedly suffer from a *mixture* of all these negative effects and thus fail to outperform local stand-alone training on clients [9, 10]. The above observations motivate the proposal of many solutions against negative impacts on FL. These solutions can be classified into two categories: (category A) global FL and (category B) personalized FL.

Research in category A aims to learn a single global model that performs uniformly well on most clients and to mitigate the negative impacts on global model learning. To do so, researchers proposed to enhance the vanilla FL paradigm with a public dataset shared among all clients for balancing their different data distributions [29], with robust aggregation designed against attacks [31], with the control variates learned collaboratively by all clients for regularizing the global model training [22–24, 32], and so on. These schemes are designed for a cooperative environment, where all clients are willing to follow the newly proposed FL algorithms that have extra computation/communication costs compared to FedAvg. Different from these schemes, ours is more flexible to allow for uncooperative clients who stick to following the vanilla FL algorithm. In a cooperative environment, though, these schemes can be combined with our solution to improve the performance of the global model.

**Table 1** Main notations used throughout this paper

| | |
|---|---|
| $N$ | Total number of clients |
| $i, r$ | Client index, round index |
| $\mathcal{D}_i$ | Training dataset on client $i$ |
| $n_i$ | Number of data samples in $\mathcal{D}_i$ |
| $\boldsymbol{w}^r$ | Global model learned by FL in round $r$ |
| $\boldsymbol{w}_i^r$ | Local update trained from $\boldsymbol{w}^{r-1}$ by client $i$ |
| $\boldsymbol{v}_i$ | Adapted model learned by FL for client $i$ |
| $V_i$ | Performance of the model produced by FL on client $i$'s testing dataset |
| $P_i$ | Performance of client $i$'s private model on its testing dataset |

Research in category B aims to learn multiple personalized (or called as adapted in this paper) models, each of which is produced for a specific client (or a subset of clients), and to mitigate the negative impacts on the model personalization/adaptation [9, 12, 18, 19, 25–28, 33, 34]. Our work is related to this category of work because the fundamental idea of our NFL recovery measures is to perform model adaptation. However, different from existing work in category B, our work not only performs model adaptation in cases of NFL but also detects NFL in run-time for evaluating the necessity of NFL recovery measures. Our proposed NFL detection scheme can be integrated with most studies in category B to save the unnecessary extra expenses that those studies introduced in the well-performing FL processes.

## 3 Negative Federated Learning

This section presents the definition of NFL. Table 1 summarizes the notations used throughout this paper. We shall start by introducing vanilla FL.

## 3.1 Preliminaries

Federated learning (FL) is a distributed learning process in which $N$ devices (i.e., clients) train a shared global model collectively without the need to centralize their private data together. Vanilla FL [1] learns the optimal global model, denoted as $w$, via iterative rounds of interactions between the clients and a central server. In each round $r$, the server first distributes the global model $w^{r-1}$ to a set of active clients, which is denoted as $C^r$. Then, each client $i \in C^r$ uses its private dataset $D_i$ to produce a locally updated model, i.e., $w_i^r \leftarrow w^r - \eta \nabla \mathcal{L}(D_i; w^r)$, and report the local update to a central server. Next, the server aggregates the received updates into a new global model via $w^r \leftarrow \sum_{i=1}^{N} \frac{n_i}{n} w_i^r$, where $n \leftarrow \sum_{i=1}^{N} n_i$. Then, the server distributes the new global model to active clients for the next round of learning.

Ideally, the performance of $w^r$ improves as the learning proceeds and approaches that of imaginary *centralized learning* (i.e., to pool all private data together and train the model on the centralized dataset). However, it turns out that FL can go wrong, resulting in most clients being unable to obtain quality models from FL and thus unwilling to contribute to FL.

## 3.2 Definition of Negative Federated Learning

To decide if FL is doing well, let us *pretend* that every client $i$ has a private model trained independently on $D_i$ beforehand. *Negative federated learning (NFL)* refers to the state of an FL system in which the model obtained from FL does not win out over the private models of *most* clients.

To formalize the definition of NFL, we first define a metric describing the performance gain obtained from FL by a client $i$. Given a model learned by FL and a private model, we denote by $V_i$ the performance of the model learned by FL, and by $P_i$ the performance of the private model, the *on-client performance gain*[2] is given by Eq. 1.

$$\beta_i \leftarrow V_i - P_i. \tag{1}$$

Next, we define a system-wide metric $\beta$, named *overall performance gain*, as following:

$$\beta \leftarrow \sum_{i=1}^{N} \alpha_i \beta_i. \tag{2}$$

Here, $\alpha_i$ is a weight indicating how much a client $i$ matters in the performance gain evaluation. Possible weight schemes could be $\alpha_i = \frac{1}{N}$ (equal weights), $\alpha_i = \frac{n_i}{n}$ (weighted by client

---

[2] Most existing performance metrics, such as accuracy or F1, can be used. To compute meaningful performance gain, we require the models in comparison to share the *same* neural structure and to be evaluated on the *same* testing dataset.

---

data size), or a positive value indicating local data quality [35].

We now define NFL as follows. For a given federated learning system (e.g., running FedAvg [1]), if there does not exist a positive integer $R$ such that for any model learned in this system after round $R$ (e.g., the global model $w^r$, where $r \geq R$, learned in the vanilla FL system), $\beta \geq 0$, then we say the system is in *negative federated learning*. In such a case, $|\beta|$ quantifies the overall negative effects on the participating clients.

Note the NFL concept presented in this paper is defined for the *system-wide* learning performance. An individual client $i$ may consider itself encountering negative learning if its $\beta_i$ remains negative and then may decide to take "individual-level measures" to improve its performance gain. We will provide a guideline for each client to perform "individual-level measures" at the end of Sect. 4.3. This issue, however, is not further discussed in other sections since our paper focuses on the collective learning behavior of the federation.

# 4 Tackling Negative Federated Learning

This section introduces a novel framework named FL-GUARD for tackling NFL in run-time of federated learning. FL-GUARD treats NFL as a *measurable, detectable, and avoidable* state that may occur in any FL process. When a system enters NFL, it can utilize our proposed approach to recover to a healthy state. The framework is named FL-GUARD because it acts like a guard of FL systems, detecting and confronting the negative learning problem. Figure 1 shows an overview of FL-GUARD. In the following subsections, we first introduce the method for detecting NFL and then detail the recovery measure.

## 4.1 Detecting Negative Federated Learning

Intuitively, NFL can be detected by checking the value of $\beta$ after each round of interaction. However, on-client model testing in each round is costly, as it incurs non-trivial extra computing on clients. To avoid such costs, we propose to use the training data as the *surrogate* to estimate the value of $\beta$ in the learning process and utilize the estimated $\beta$ value for NFL detection.

We take the vanilla FL system as an example to illustrate how $\beta$ is estimated efficiently during the learning run-time. In each round $r$, each participating client $i$ first utilizes its first training batch $b_i^r$ to estimate its performance gain from engaging in FL by Eq. 3, where $EV$ is a function for evaluating model performance on a given dataset, $w^{r-1}$ is the latest global model that client $i$ received from the vanilla FL system, and $P_i$ is the performance of client $i$'s private model on its testing dataset. The private model that client $i$ possessed before engaging in FL is not updated in the FL run-time. Thus, client $i$ only needs to compute the value of $P_i$ once before FL starts.

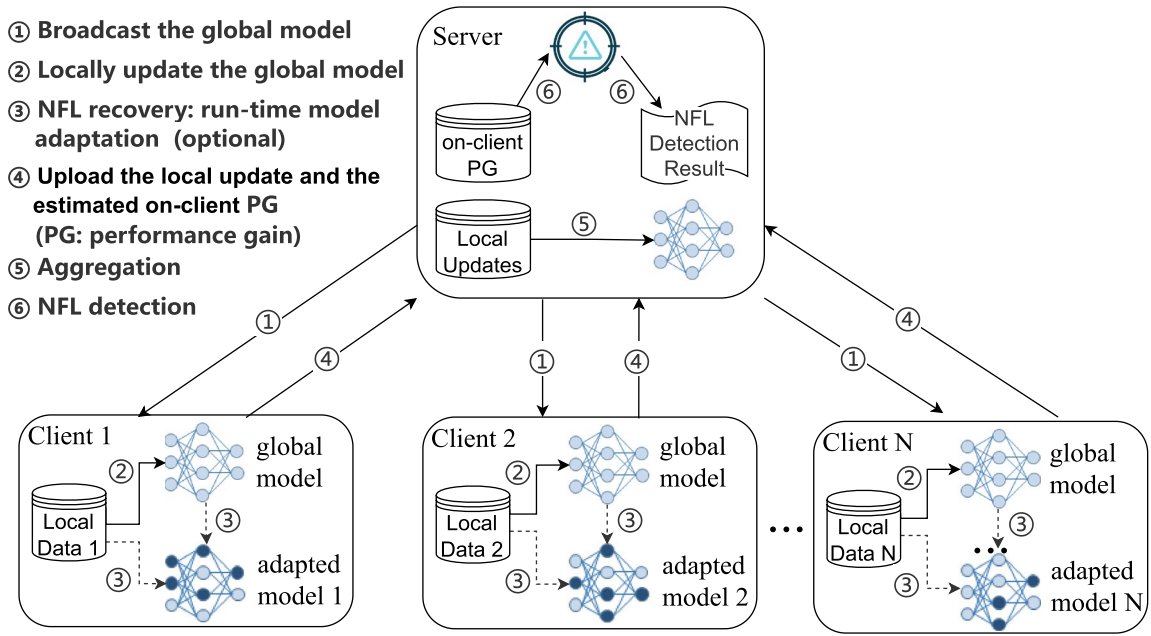

**Fig. 1** An overview of FL-GUARD. *Dashed arrows* denote the optional model adaptation, which is activated after NFL is detected, for recovering to a healthy learning state

$$\hat{\beta}_i \leftarrow EV(\boldsymbol{w}^{r-1}, b_i^r) - P_i. \qquad (3)$$

Note that client $i$ can easily obtain the value of $EV(\boldsymbol{w}^{r-1}, b_i^r)$ when performing the feed-forward training of model $\boldsymbol{w}^{r-1}$. Thus, the additional computation cost introduced by computing Eq. 3 is trivial. $EV(\boldsymbol{w}^{r-1}, b_i^r)$ in Eq. 3 is an unbiased estimate of the performance of model $\boldsymbol{w}^{r-1}$ on client $i$'s testing dataset (i.e., $V_i$ in Eq. 1). This is because the batch of data $b_i$ is randomly sampled from client $i$'s dataset $D_i$ and has a similar distribution to that of client $i$'s complete testing dataset. When submitting its local update, client $i$ reports the computed $\hat{\beta}_i$ value to the server.

Given all $\hat{\beta}_i$ values reported from clients $i \in C^r$ in round $r$, the overall performance gain $\hat{\beta}$ is estimated on the server by two steps:

(1) The server computes a $\hat{\beta}^r$ value by Eq. 4 which selects the median of $\{\hat{\beta}_i | i \in C^r\}$.

$$\hat{\beta}^r \leftarrow Median(\{\hat{\beta}_i | i \in C^r\}). \qquad (4)$$

(2) The server computes the overall performance gain $\hat{\beta}$ by Eq. 5 which averages the $\hat{\beta}^r$ values computed in the last $c$ rounds.

$$\hat{\beta} \leftarrow \frac{1}{c} \sum_{c'=0}^{c-1} \hat{\beta}^{r-c'}. \qquad (5)$$

The above two steps are designed under the consideration of malicious clients who may report the fabricated $\hat{\beta}_i$ and lead to undesirable fluctuation of $\hat{\beta}$. With these two steps, a malicious client can hardly influence the estimated overall performance gain, no matter whether its fabricated $\hat{\beta}_i$ is close to or diverging from the ones submitted by honest clients.

The server uses $\hat{\beta}$ to detect the NFL state. A negative $\hat{\beta}$ indicates that the model learned by FL performs worse than most clients' private models and implies the state of NFL. An advantage of using $\hat{\beta}$ is that it can be obtained without the high cost of evaluating the learned model on (large) datasets in each round.

**NFL Detection.** Based on the above discussion, we can develop a cost-effective NFL detection scheme on the server as follows:

(1) During iterative FL, the server monitors if $\hat{\beta} < 0$ after each round of client–server interactions. If it is true in more than $NR$ rounds, where $NR$ is a threshold for Negative Rounds, the server reports a state of NFL.

(2) A reported NFL can be *canceled* as well. If $\hat{\beta} \geq 0$ is observed for $c$ consecutive rounds and NFL has previously been reported, the server cancels the NFL report, but the detection process continues.

For early detection of NFL (e.g., within tens rounds), *NR* is recommended to be less than 100. But too small *NR* value may produce a false positive (i.e., to report well-performing FL as NFL). Fortunately, this does not negatively influence the learning performance as it can be canceled in subsequent rounds when $\hat{\beta} \geq 0$ is observed for *c* consecutive rounds. False positives only incur some (unnecessary) recovery steps before the NFL report is canceled.

Once NFL is detected in a system, it is necessary to take some measures to improve the performance of the model learned by FL on individual clients. We achieve this by making local adaptation to the global model learned by the vanilla FL paradigm when that global model is iteratively trained on each client. The details of model adaptation are presented in the next section.

### 4.2 Recovering From Negative Federated Learning

This section presents how we perform *model adaptation* on the global model learned by vanilla FL when that global model cannot perform well on most clients. While it might not be impossible to learn a single global model that performs uniformly well on most (or even all) clients, we believe that performing model adaptation in FL run-time is a much easier approach for achieving better *local performance*, especially when some clients may fail in taking any NFL recovery measures or deliberately disrupt the learning system.

Our model adaptation approach introduces for each client an *adapted model* ($v_i$), which can fit the local data better and is tightly coupled in the process of learning the *global model* ($w$). In each round *r*, model adaptation updates the adapted model ($v_i$) and the global model ($w^{r-1}$) *simultaneously* on client *i*. The objective function for optimizing the global model is given by $\ell(w^{r-1}, \mathcal{D}_i)$. We do not explain this objective function in detail as it is the same as that in the vanilla FL system.

To optimize the adapted model $v_i$, we minimize two terms: (1) the loss of the adapted model on local training data $\ell(v_i, \mathcal{D}_i)$ and (2) the divergence in the parameters of the adapted model and the learned global model $\|v_i - w_i^r\|^2$ (where $w_i^r$ is initialized by $w^{r-1}$ in each round). The first term is minimized for making the adapted model fit the local data, while the second is minimized for improving the adapted model's generalization performance. We empirically find that indiscriminately minimizing the second term may not always help. When the global model performs well on most clients, minimizing $\|v_i - w_i^r\|^2$ on those clients can improve the generalization performance of their adapted models. In contrast, if the global model performs much worse than the local optimal model on a client or diverges significantly from that client's local optimal model, minimizing $\|v_i - w_i^r\|^2$ on that client may worsen the performance of its adapted model. We introduce a parameter $\lambda$ to *flexibly* control the minimization of $\|v_i - w_i^r\|^2$. $\lambda$ is computed by Eq. 6–8, where $\sigma(\cdot)$ is the sigmoid function and $<,>$ is the dot product of two vectors. With $\lambda$, the objective function for optimizing the adapted model is given by Eq. 9.

$$loss\_div \leftarrow \ell(v_i, \mathcal{D}_i) - \ell(w_i^r, \mathcal{D}_i). \tag{6}$$

$$grad\_div \leftarrow \frac{< v_i - w_i^r, \nabla_{v_i}\ell(v_i, \mathcal{D}_i) >}{\|\nabla_{v_i}\ell(v_i, \mathcal{D}_i)\|}. \tag{7}$$

$$\lambda \leftarrow \sigma(loss\_div) \times \sigma(grad\_div). \tag{8}$$

$$\mathcal{L}_a \leftarrow \ell(v_i, \mathcal{D}_i) + \lambda\|v_i - w_i^r\|^2. \tag{9}$$

The value of $\lambda$ would become small as the result of two cases. First, the global model performs poorly, causing a large loss $\ell(w_i^r, \mathcal{D}_i)$ and thus the first $\sigma(\cdot)$ in Eq. 8 gives a small value. Second, the parametric difference ($v_i - w_i^r$) strongly disagrees with the gradient of $\ell(v_i, \mathcal{D}_i)$, so the second $\sigma(\cdot)$ produces a small output. In both cases, we want to play down the impacts of the global model and allow $\ell(v_i, \mathcal{D}_i)$ to dominate the learning process. Whereas if the global model has a much smaller training error and there is an agreement between the two minimization goals (i.e., to minimize $\ell(v_i, \mathcal{D}_i)$ or $\|v_i - w_i^r\|^2$), it is relatively safer to put more emphasis on the second goal. The above discussion explains the usage of $\lambda$ in Eq. 9.

**Algorithm 1** FL-GUARD

---

**Input** : System parameters $N, R, B, E, \eta, c, NR$;
The indicator of NFL, $\mathbb{I}_{NFL}$.

**Server executes:**

s1  initialize the indicator of NFL by
$\mathbb{I}_{NFL} \leftarrow False$

s2  initialize the global model $\boldsymbol{w}^0$ and an
integer $cnt \leftarrow 0$

s3  **for** *each round r=1,2,..., R* **do**

s4      **ask** *client $i \in C^r$ in parallel* **do**

s5        $\boldsymbol{w}_i^r, \hat{\beta}_i^r \leftarrow$ ClientUpdate($i$,
$\boldsymbol{w}^{r-1}, \mathbb{I}_{NFL}$)

s6      $\boldsymbol{w}^r \leftarrow \sum_{i=1}^{N} \frac{n_i}{\sum_{i=1}^{N} n_i} \boldsymbol{w}_i^r$

s7      $\hat{\beta}^r \leftarrow Median(\{\hat{\beta}_i | i \in C^r\})$

s8      $\hat{\beta} \leftarrow \frac{1}{c} \sum_{c'=0}^{c-1} \hat{\beta}^{r-c'}$

s9      **if** $\hat{\beta} < 0$ **then** $cnt \leftarrow cnt + 1$

s10     **if** $cnt > NR$ & $\neg \mathbb{I}_{NFL}$ **then**

s11       report NFL by setting
$\mathbb{I}_{NFL} \leftarrow True$

s12     $\epsilon \leftarrow \max(\{r' | r' \in [1, r] \wedge \hat{\beta}^{r'} < 0\})$

s13     **if** $\mathbb{I}_{NFL}$ & $r - \epsilon > c$ **then**

s14       cancel NFL report by setting
$\mathbb{I}_{NFL} \leftarrow False$

**Client executes:**

c1  **ClientUpdate($i, \boldsymbol{w}^{r-1}, \mathbb{I}_{NFL}$):**

c2    $\mathcal{B} \leftarrow$ split local data $D_i$ into
batches of size $B$

c3    **if** $\mathbb{I}_{NFL}$ & $\boldsymbol{v}_i$ *not initialize* **then**

c4      initialize $\boldsymbol{v}_i$

c5    **if** $\neg \mathbb{I}_{NFL}$ **then**

c6      compute $\hat{\beta}_i$ by Eq. 3

c7    **else**

    $\hat{\beta}_i \leftarrow EV(\boldsymbol{v}_i, b_i^r) - P_i$

c8    $\boldsymbol{w}_i^r \leftarrow \boldsymbol{w}^{r-1}$

c9    **for** *each epoch e = 1, 2, ..., E* **do**

c10     **for** *each batch $b \in \mathcal{B}$* **do**

c11       $\boldsymbol{w}_i^r \leftarrow \boldsymbol{w}_i^r - \eta \nabla_{\boldsymbol{w}_i^r} \ell(\boldsymbol{w}_i^r, b)$

c12       **if** $\mathbb{I}_{NFL}$ **then**

c13         compute $\lambda$ by Eq. 6–8

c14         compute $\mathcal{L}_a$ by Eq. 9

c15         $\boldsymbol{v}_i \leftarrow \boldsymbol{v}_i - \eta \nabla_{\boldsymbol{v}_i} \mathcal{L}_a$

c16   **upload** ($\boldsymbol{w}_i^r, \hat{\beta}_i$) to server

---

### 4.3 FL-GUARD Operations

Algorithm 1 presents the complete pseudo-code for implementing our FL-GUARD framework, which contains *NFL detection* and *the model adaptation for NFL recovery*, upon the vanilla FL system. At the end of each round $r$, clients return only the local update $\boldsymbol{w}_i^r$ along with a float value $\hat{\beta}_i$ to the server. Thus, the communication cost of our FL-GUARD is almost the same as that of vanilla FL. Moreover, it is noticeable that the transmission of $\hat{\beta}_i$ would not incur new challenges in privacy protection like the extra variates transmitted in previous NFL solutions, e.g., SCAFFOLD [22], FedNova [23], FedLin [24], etc.

There are two modes to use FL-GUARD: (1) *Detection and recovery*, which is described in Algorithm 1 and highly recommended. An FL system running in this mode can operate without introducing the costs for NFL recovery until it detects NFL. Only when NFL is reported, an indicator denoted as $\mathbb{I}_{NFL}$ is set. Then, the clients start recovery by activating model adaptation. (2) *All-time recovery*. Alternatively, the system may choose to activate NFL recovery in the entire learning life cycle by setting $\mathbb{I}_{NFL} \leftarrow True$ at line

s1 and canceling operations at lines s7–s14. For any mode, activating the model adaptation will not change the results of learning the global model. However, once model adaptation is activated, each client utilizes the adapted model (instead of the learned global model) for its local inference/testing. Thus, the performance gain metric ($\beta_i$) is estimated as $\hat{\beta}_i \leftarrow EV(\boldsymbol{v}_i, b_i^r) - P_i$ for client $i$. We recommended the "detection and recovery" mode, since our NFL detection is inexpensive and the costs for (unnecessary) NFL recovery can be avoided in a well-performing FL process.

As an additional refinement, we provide a guideline for each client to perform "individual-level measures" in FL-GUARD. Specifically, each client can independently start model adaptation when the system-wide NFL state is not reported, if its locally computed $\hat{\beta}_i$ remains negative in more than $NR$ rounds. Moreover, each client can stop its activated model adaptation to save the computation cost, if it observed $\hat{\beta}_i \geq 0$ for $c$ consecutive rounds. FL-GUARD allows these "individual-level measures" to be performed because these measures will not pose harmful impacts on the system-wide learning results.

**Table 2** Task profiles and default experimental settings

| | CIFAR | SHAKE |
|---|---|---|
| Dataset (# Classes) | CIFAR-10 (10) | Shakespeare (80) |
| Neural model | CNN | LSTM |
| # Clients ($N$) | 100 | 100 |
| Data allocation | non-IID (mixed) | non-IID (by role) |
| % Active clients | 10% | 10% |
| % Attackers | 30% | 30% |
| $\mathcal{S}$ in diff. privacy | 15 | 50 |
| $\sigma$ in diff. privacy | 0.001 | 0.001 |
| Optimizer | SGD | SGD |
| Learning rate ($\eta$) | 0.1 | 1.47 |
| # Rounds ($R$) | 1000 | 500 |
| Local epochs ($E$) | 1 | 1 |
| Local batch size ($B$) | 10 | 32 |

The architecture of the neural model trained in each FL task is the same as described in previous paper [1]

### 4.4 Advantage of FL-GUARD

Compared to existing solutions for tackling NFL, FL-GUARD has the following advantages:

(1) FL-GUARD tackles NFL dynamically. It can not only activate recovery in system run-time based on the result of NFL detection but also cancel the NFL report and stop recovery when it becomes unnecessary. Such *dynamicity* avoids the extra expenses in well-performing FL processes. More importantly, as we will see in the experiment, when vanilla FL is sufficient to perform well, indiscriminate activation of NFL recovery measures (no matter ours or most of the previous ones) may slightly *harm* the performance. This result justifies the need for the detection and recovery mode.

(2) FL-GUARD can be easily employed upon any FL system for run-time NFL detection and recovery and is compatible with most existing NFL handling techniques. This advantage is demonstrated empirically in Sect. 5.4. An important benefit of integrating existing NFL recovery techniques with our NFL detection scheme is to save (unnecessary) extra expenses of these techniques when vanilla FL is sufficient to perform well.

(3) FL-GUARD is robust against clients who do not take any NFL recovery measures. As mentioned in Sect. 1, when upgrading an FL system in use, it is common to see (i) some clients unwilling to adopt the newly proposed techniques until observing the others benefit from doing so and (ii) some clients unable to adopt the newly proposed techniques due to their limited computing/communicating power. These clients tend to remain as *vanilla FL clients*, who stick to following FedAvg and intentionally disable all NFL recovery measures even when the FL system is in NFL state. Fortunately, our *FL-GUARD clients* (i.e., those activating model adaptation when NFL is detected) can always achieve performance gains from FL, even if the others (e.g., attackers, resource-constrained clients, etc.) remain to be vanilla FL clients. Such robustness, as shown in Sect. 5.4, is important for employing a newly proposed technique to upgrade an FL system in use.

## 5 Experiments and Results

We evaluate the performance of FL-GUARD on two typical federated learning tasks, namely CIFAR (*image classification* on CIFAR-10 [36]) and SHAKE (*language modeling* on Shakespeare dataset [7]). The major issues that may lead to NFL, which include data heterogeneity among clients, client inactivity, attacks from malicious clients, and noises introduced by differential privacy protection, are all considered in our FL tasks.

Table 2 summarizes the default environment settings in different tasks and the hyperparameters used for model training. Our experiment settings all follow previous literature. Unless otherwise stated, all experiments in this paper are conducted under the default settings specified in Table 2. Below, we detail the default environment setup.

**Simulation of non-IID local data.** We simulate the unbalanced and non-IID data distributions across clients in a way similar to that of previous work [1, 11]. For CIFAR-10, we allocate its 50,000 training samples and 10,000 testing samples to 100 clients based on a "*non-IID (mixed)*" scheme. Specifically, we first sort the CIFAR-10 dataset by labels and then set up a case where 50 clients have samples from 10 classes, 30 clients have samples from 5 classes, and the remaining 20 clients have samples from 2 classes. Moreover, we set the amount of data allocated to clients following a *Lognormal*$(0, 2^2)$ distribution.

For Shakespeare, we follow previous work [1, 7, 37] to allocate each speaking role to one client and term this allocation scheme as *"non-IID (by role)"*. Following Wang et al. [37], we filter out the clients with less than 10,000 data samples and then sample a random subset of 100 clients. In each client, we allocate 90% of the data for local training and the remaining for testing. The sampled dataset contains a total of 1609724 samples for training and 171017 for testing.

We also form a balanced and *IID* data allocation on CIFAR-10 and Shakespeare for reference and comparison. In the *IID* scheme, the original CIFAR-10/Shakespeare dataset is first shuffled and then allocated to 100 clients. Each

**Fig. 2** Using $\hat{\beta}$ for NFL detection. Results are reported based on the vanilla FL system. The pink curve shows results under the default environment setup specified in Table 2. The blue curve shows results under an *Ideal FL* scenario, in which we remove most of the negative effects on FL by using IID data allocations in both tasks and removing all attackers

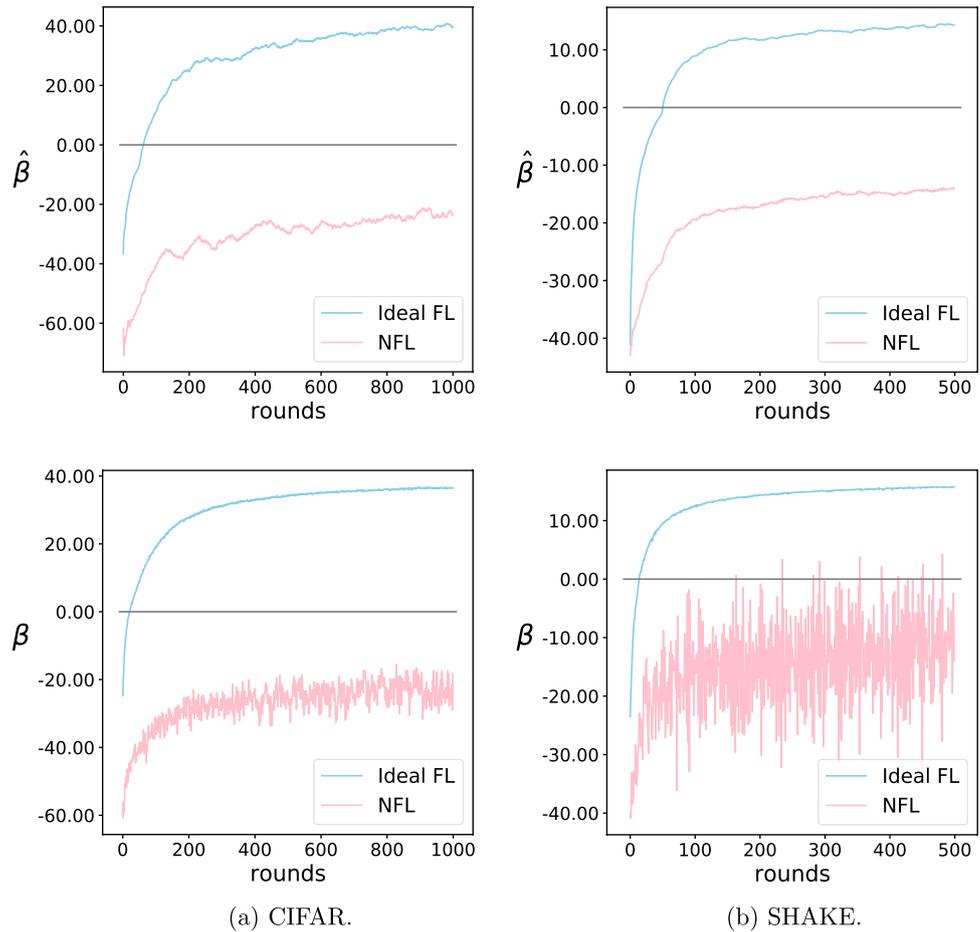

(a) CIFAR.                    (b) SHAKE.

**Table 3** Tuning the parameter *NR* used in the NFL detection scheme

| NR | NFL | | Ideal FL | |
|---|---|---|---|---|
| | CIFAR | SHAKE | CIFAR | SHAKE |
| 10 | 10 | 10 | 10 | 10 |
| 30 | 30 | 30 | 37 | 30 |
| 50 | 50 | 50 | – | 50 |
| 70 | 70 | 70 | – | – |
| 90 | 90 | 90 | – | – |

We show the index of the round in which NFL is reported. The bar (–) means that NFL is not reported. We recommend *NR* = 50 and *NR*= 70 to be set, respectively, on CIFAR and SHAKE for both a quick response to NFL under the default setup specified in Table 2 and no false positive under the ideal FL setup specified in Fig. 2 client receives the *same* amount of data samples and owns the samples from all ten classes in CIFAR-10 and all 80 classes in Shakespeare.

**Simulation of client inactivity.** We simulate client inactivity by uniformly random sampling 10% of all clients ($K/N = 10\%$, where $K$ is the number of active clients) in every round to participate in model training, which is the same as the setting in most of the previous work [1, 10, 23, 38, 39]. Among the selected clients in every round, several clients are malicious, poisoning the global model being learned by FL. The details of attacks are given in the next paragraph.

**Simulation of attacks.** Our experiments randomly select some of the clients to be malicious attackers who poison the global model via reporting the model parameters updated on label-flipped data samples. Details for attacking strategy can be found in previous work [40]. Based on the state-of-the-art attacks on FL [40–42], we set 30% of all clients to be attackers.

**Simulation of differential privacy protection.** We follow previous work [10, 16] to simulate the differential privacy protection on the server, i.e., to aggregate local updates and produce the updated global model in each round $r$ by $w^r \leftarrow w^{r-1} + \frac{1}{K} \sum_{i \in C^r} Clip(w_i^r - w^{r-1}, S) + \mathcal{N}(\mathbf{0}, \sigma^2 \mathbf{I})$, where the norm of each local update is clipped with an upper bound $S$ and Gaussian noise $\mathcal{N}(\mathbf{0}, \sigma^2 \mathbf{I})$ is added. Hyper-parameters are set as $S = 15$, $\sigma = 0.001$ for CIFAR, and $S = 50$, $\sigma = 0.001$ for SHAKE.

We mainly monitor two metrics to evaluate whether FL brings benefits to its participating clients: (1) *the average*

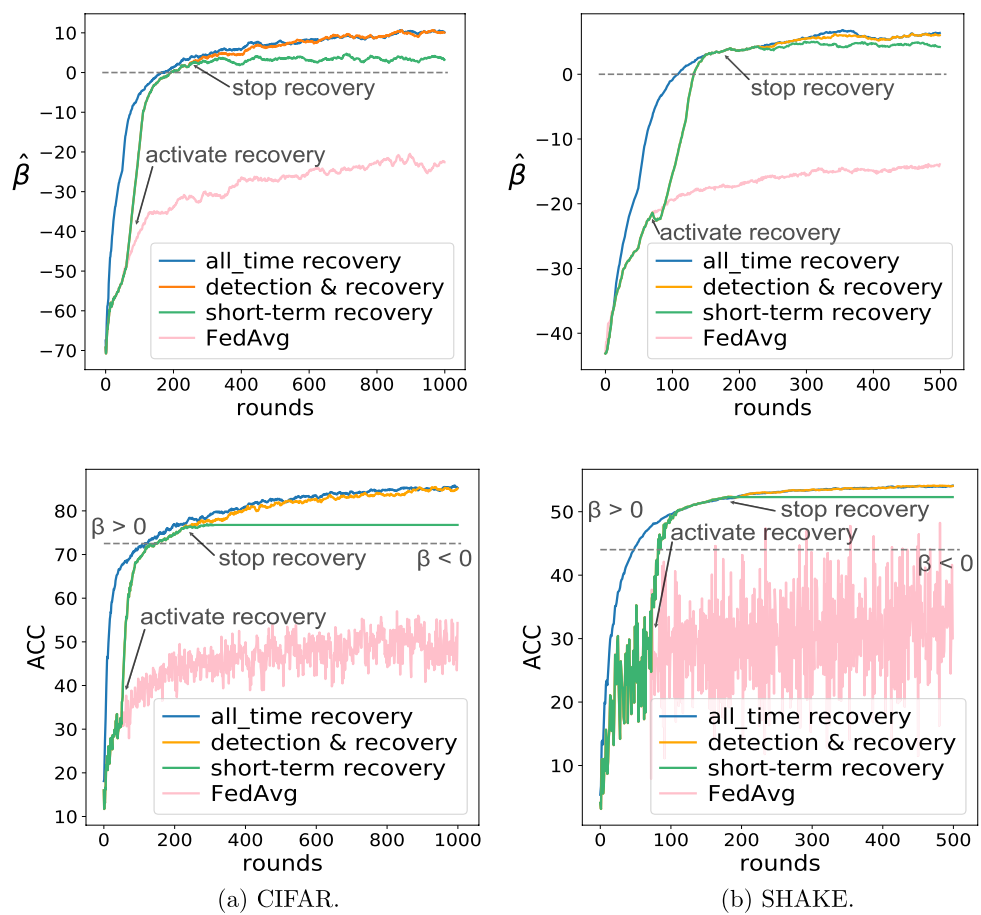

**Fig. 3** Run-time results of FL-GUARD under the default environment settings. The gray lines in the figures of ACC indicate the performance of local stand-alone training obtained before FL starts

(a) CIFAR.      (b) SHAKE.

*local accuracy (ACC)*, evaluated for the model that the FL system provides for on-client local inference by testing that model on each client's private testing dataset; and (2) *the average performance gain* ($\beta$), compared to local stand-alone training[3].

Each FL process is given fixed rounds of learning (i.e., 1000 rounds for CIFAR and 500 rounds for SHAKE). All experiments are repeated for three runs with different random seeds. We report the averaged results over these runs, and the results shown in the tables are averaged over the last 10 rounds.

### 5.1 Results of NFL Detection and Recovery

We first justify the usefulness of $\hat{\beta}$ for NFL detection and then study the proposed NFL detection/recovery scheme.

**Using $\hat{\beta}$ for NFL Detection.** Fig. 2 reports the results of $\hat{\beta}$ when the vanilla FL system is in NFL and ideal FL, respectively. We also plot each round's true $\beta$ value for reference. In the NFL state (pink curve), both $\hat{\beta}$ and true $\beta$ value remain fluctuating below zero, implying the poor performance of the federated learning on individual clients. In contrast, the value of $\hat{\beta}/\beta$ quickly goes above zero in ideal FL. $\hat{\beta}$ tells NFL from ideal FL in only tens of learning rounds. The results confirm the usefulness of $\hat{\beta}$ as a surrogate of true $\beta$ for fast NFL detection. Note that the amplitude of fluctuation of $\hat{\beta}$ is smaller than that of $\beta$ in NFL. This is mainly due to the smoothing effect of averaging $\hat{\beta}$ over the last $c = 50$ rounds.

---
[3] To calculate $\beta$, we let $\alpha_i = \frac{1}{N}$ in Eq. 2. Other weights result in the same conclusions. The accuracy of local stand-alone training is omitted as it is equal to the reported *ACC* minus $\beta$.

**The *NR* Threshold for NFL Detection.** Table 3 presents the number of learning rounds needed by our detection scheme, with different *NR* (Negative Rounds) thresholds, for reporting NFL. Generally, our detection method is very quick in responding to NFL, though it causes false positives in the case of ideal FL when *NR* < 50 on CIFAR and *NR* < 70 on SHAKE (since the threshold value is too small). As *NR* increases, false positives disappear, but it takes more rounds for the system to report NFL. False positives have negligible impacts on learning performance (since they can be canceled when $\hat{\beta} \geq 0$ is observed for $c$ consecutive rounds). However, they (unnecessarily) incur double learning costs on all clients. Therefore, we set a moderate value of *NR* = 50 for CIFAR and *NR* = 70 for SHAKE. Learning on SHAKE is slower in convergence speed than learning on CIFAR and thus needs a greater *NR* to prevent false positives.

**Learning with NFL detection and recovery.** With the above recommended parameter setting, we monitor $\hat{\beta}$ and run-time ACC of FL-GUARD running in both *detection and recovery* and *all-time recovery* modes. To better study the effect of recovery, we set the recovery in the above modes as long-term (i.e., never stopped after being activated) and try an additional *short-term recovery* mode, where recovery is activated by NFL detection, and then after $\hat{\beta} > 0$ is observed in consecutive $c = 50$ rounds, the system cancels NFL report and stops recovery. Note that *short-term recovery* is precisely the same as *detection and recovery* before stopping recovery.

The results shown in Fig. 3 reveal the following findings:

- First, the detection scheme reports NFL quickly (in tens of learning rounds), confirming the usefulness of $\hat{\beta}$ for fast NFL detection.
- Next, as a result of activating recovery, the green ACC increases rapidly, overtaking the gray line (performance of local stand-along learning), and soon approaches the blue ACC (*all-time recovery*). These results indicate the effectiveness of model adaptation in run-time NFL recovery.
- The results of the green ACC also indicate that once recovery stops, the performance stops growing too. However, the system does not return to NFL because the adapted models are kept by the clients for local inference.
- Finally, the orange curves show further performance improvements if the recovery continues, getting almost the same final ACCs as the *all-time recovery*.

All the above findings clearly show the effectiveness and efficiency of our proposed detection and recovery scheme and the performance improvements from model adaptation.

## 5.2 Comparison with Previous Methods

We compare FL-GUARD with the following approaches:

1. *FedProx* [38], to utilize a proxy regularization term for tackling data statistical heterogeneity in the federation;
2. *FedMedian* [31], robust aggregation approach which takes the coordinate-wise median of local updates for preventing the global model from being poisoned;
3. *TrimmedMean* [31], similar to FedMedian, but takes the coordinate-wise trimmed mean of local updates in aggregation;
4. *multi-Krum* [43], another robust aggregation approach which discards the top-*k* local updates with a relatively large distance to the others;

5. *K-norm* [28], similar to multi-Krum, but discards the top-*k* local updates with relatively large norms;
6. *FT* [12], to fine-tune the global model on the client after an FL process ends;
7. *FB* [12], similar to FT, but fine-tunes only the top layer of the global model on the client after an FL process ends;
8. *KD* [10], to augment FT with knowledge distillation;
9. *MTL* [10], to augment FT with multitask learning;
10. *PerFedAvg(HF)* [44], similar to FT, but utilizes the meta-learning technique for training the global model;
11. *APFL* [21], a recent adaptation approach to integrate the global model with a per-client local model;
12. *Ditto* [28], a recent work similar to ours, but regularizes the adapted model with a frozen copy of the global model.

For a fair comparison, all the above approaches are implemented upon the vanilla FL (FedAvg) system. The results of FedAvg are also reported for reference. Our NFL solution is compared with only the most competitive, instead of all, work in each category of existing NFL handling approaches (presented in Sect. 2) because the performance of the approaches in the same category is similar. Later in Sect. 5.4, we will show that our NFL solution is compatible with many existing NFL handling approaches.

**Comparison under the default environment settings.** As presented in the upper half of Table 4, FL-GUARD outperforms all previous approaches in tackling NFL in the vanilla FL system, with its *all-time recovery* mode leading on all metrics. When the *detection and recovery* mode is used, the performance of FL-GUARD slightly drops ($< 0.8$), but it still outperforms most previous approaches.

**Comparison under the *Ideal FL* environment settings.** The lower half of Table 4 presents the comparative results under the ideal FL environment settings. Almost all previous approaches cannot compete with the vanilla FL paradigm (FedAvg). Some even cause a reduction of ACC as much as 6%, and FL-GUARD with all-time recovery also harms learning performance. In contrast, FL-GUARD running in the *detection and recovery* mode still performs better on both datasets compared to the other methods. The reason is that, in the detection and recovery mode, FL-GUARD never activates recovery when FL performs well without it. In this case, FL-GUARD is exactly the same as vanilla FL! This explains the identical results of FedAvg and the bottom row and justifies the need for *detection and recovery*.

The above results reveal the most important advantage of FL-GUARD: the ability to dynamically tackle NFL in run-time, a feature not reported in any previous work. If NFL occurs, it can be detected and recovered. Whereas if NFL

**Table 4** Comparison with previous FL methods

| Metric | CIFAR | | SHAKE | |
|---|---|---|---|---|
| | ACC | $\beta$ | ACC | $\beta$ |
| **Default Settings** | | | | |
| *Vanilla FL (FedAvg)* | 48.26 | − 24.26 | 33.74 | − 10.28 |
| *FedProx* | 49.94 | − 22.58 | 29.02 | − 15.00 |
| *FedMedian* | 26.25 | − 46.27 | 40.80 | − 3.22 |
| *TrimmedMean* | 53.03 | − 19.49 | 45.14 | +1.12 |
| *multi-Krum* | 55.42 | − 17.10 | 48.81 | +4.79 |
| *K-norm* | 55.95 | − 16.57 | 48.18 | +4.16 |
| *FT* | 82.86 | +10.34 | 52.70 | +8.68 |
| *FB* | 83.98 | +11.46 | 53.68 | +9.66 |
| *KD* | 83.20 | +10.68 | 52.74 | +8.72 |
| *MTL* | 82.39 | +9.87 | 51.98 | +7.96 |
| *PerFedAvg(HF)* | 83.30 | +10.78 | 52.51 | +8.49 |
| *APFL* | 83.93 | +11.41 | 53.99 | +9.97 |
| *Ditto* | 85.21 | +12.69 | 52.75 | +8.73 |
| *FL-GUARD* | | | | |
| All-time recov | **85.53** | **+13.01** | **54.04** | **+10.02** |
| Detect & recov | 84.85 | +12.33 | 54.03 | +10.01 |
| **Ideal FL Settings** | | | | |
| *Vanilla FL (FedAvg)* | 80.27 | +36.57 | 58.20 | +15.75 |
| *FedProx* | 80.20 | +36.50 | 58.16 | +15.72 |
| *FedMedian* | 57.18 | +13.48 | 48.74 | +6.28 |
| *TrimmedMean* | 79.01 | +35.31 | 58.06 | +15.61 |
| *multi-Krum* | 80.05 | +36.35 | 58.13 | +15.68 |
| *K-norm* | 80.06 | +36.36 | 58.12 | +15.67 |
| *FT* | 73.52 | +29.82 | 52.52 | +10.07 |
| *FB* | 76.45 | +32.75 | 54.15 | +11.70 |
| *KD* | 73.80 | +30.10 | 52.95 | +10.5 |
| *MTL* | 77.11 | +33.41 | 55.86 | +13.41 |
| *PerFedAvg(HF)* | 73.62 | +29.92 | 52.71 | +10.26 |
| *APFL* | 76.59 | +32.89 | 55.32 | +12.87 |
| *Ditto* | 78.18 | +34.48 | 57.05 | +14.60 |
| *FL-GUARD* | | | | |
| All-time recov | 77.48 | +33.78 | 56.95 | +14.50 |
| Detect & recov | **80.27** | **+36.57** | **58.20** | **+15.75** |

FL-GUARD with its detection and recovery mode can outperform most previous approaches under both the default (NFL) settings and the ideal FL settings

Bold values indicate the better performance of FL-GUARD (our approach) compared with other baselines under each experimental setting never occurs, the system proceeds based on vanilla FL without taking any (unnecessary) compensation measures. In the rest of the experiments, we shall use *detection and recovery* as the default operation mode of FL-GUARD.

## 5.3 Tuning Environment Settings

We further tune the federated environment settings to study their impacts on the performance of FL-GUARD. Since this paper employs FL-GUARD upon the vanilla FL system (FedAvg), we also report the results of *FedAvg* for reference. In each test, we vary one environmental parameter only while keeping the rest as their default values. The asterisks (*) indicate the default values in our experiments.

Tables 5, 6, 7 and 8 show the tuning results. Generally, NFL is prevalent in the entire parameter space that we tune. In most cases, FedAvg can hardly reach high accuracy. In contrast, FL-GUARD can always ensure a positive gain in accuracy ($\beta > 0$) and in many cases $\beta > 10$. Such a gain is important, as it is the main incentive for clients to participate in FL.

Table 5 shows the impacts of local data distributions on FL. When data distributions differ among clients, a significant reduction is evident in the results of FedAvg reported on CIFAR. In contrast, FL-GUARD maintains a high learning accuracy, which confirms its effectiveness in making a desirable local adaptation on the client data. The non-IID data allocation schemes in both CIFAR and SHAKE can make local data fitting much easier. Thus, on each client, the model adaptation and local stand-alone training both perform much better, with the latter improved much more. This explains the higher *ACC* but reduced $\beta$ achieved by FL-GUARD under the non-IID local data. Compared to the results reported on CIFAR, the impacts of data distributions on FL are not significant in the results of SHAKE. This is mainly because the non-IID allocation scheme of SHAKE does not significantly enlarge the difference in client data distributions (e.g., the number of class labels allocated to each client is almost the same

**Table 5** Varying data distributions among clients

| Data Alloc | ACC | | $\beta$ | |
|---|---|---|---|---|
| | FedAvg | FL-GUARD | FedAvg | FL-GUARD |
| (CIFAR) | | | | |
| IID | 76.69 | **76.69** | +32.87 | **+32.87** |
| Non-IID* | 48.26 | **84.85** | − 24.26 | **+12.33** |
| (SHAKE) | | | | |
| IID | 32.50 | **52.87** | − 9.95 | **+10.42** |
| Non-IID* | 33.74 | **54.03** | − 10.28 | **+10.01** |

FL-GUARD can maintain the positive performance gain in accuracy when data distributions differ among clients

Bold values indicate the better performance of FL-GUARD (our approach) compared with other baselines under each experimental setting

**Table 6** Varying the ratio of active clients in each round

| K/N | ACC | | β | |
|---|---|---|---|---|
| | FedAvg | FL-GUARD | FedAvg | FL-GUARD |
| (CIFAR) | | | | |
| 90% | 57.23 | **84.98** | − 15.29 | **+12.46** |
| 50% | 57.41 | **84.86** | − 15.11 | **+12.34** |
| 10%* | 48.26 | **84.85** | − 24.26 | **+12.33** |
| (SHAKE) | | | | |
| 90% | 35.05 | **54.67** | − 8.97 | **+10.65** |
| 50% | 34.65 | **54.23** | − 9.37 | **+10.21** |
| 10%* | 33.74 | **54.03** | − 10.28 | **+10.01** |

FL-GUARD is robust against the negative effects introduced by the decreasing ratio of active clients

Bold values indicate the better performance of FL-GUARD (our approach) compared with other baselines under each experimental setting

**Table 7** Varying the proportion of attackers in each round

| Attack | ACC | | β | |
|---|---|---|---|---|
| | FedAvg | FL-GUARD | FedAvg | FL-GUARD |
| (CIFAR) | | | | |
| 0% | 74.18 | **86.44** | +0.98 | **+13.24** |
| 10% | 72.09 | **86.35** | − 1.00 | **+13.26** |
| 30%* | 48.26 | **84.85** | − 24.26 | **+12.33** |
| (SHAKE) | | | | |
| 0% | 58.30 | **58.30** | +13.86 | **+13.86** |
| 10% | 57.52 | **57.52** | +13.18 | **+13.18** |
| 30%* | 33.74 | **54.03** | − 10.28 | **+10.01** |

FL-GUARD is resilient against the negative effects introduced by attackers

Bold values indicate the better performance of FL-GUARD (our approach) compared with other baselines under each experimental setting

**Table 8** Varying the std (σ) of noises introduced by differential privacy

| σ | ACC | | β | |
|---|---|---|---|---|
| | FedAvg | FL-GUARD | FedAvg | FL-GUARD |
| (CIFAR) | | | | |
| 0.001* | 48.26 | **84.85** | − 24.26 | **+12.33** |
| 0.003 | 43.68 | **83.05** | − 28.84 | **+10.52** |
| 0.005 | 38.91 | **78.23** | − 33.61 | **+5.71** |
| (SHAKE) | | | | |
| 0.001* | 33.74 | **54.03** | − 10.28 | **+10.01** |
| 0.003 | 32.98 | **54.03** | − 11.04 | **+10.01** |
| 0.005 | 33.05 | **53.94** | − 10.97 | **+9.92** |

FL-GUARD is resilient against the negative effects introduced by privacy protection measures

Bold values indicate the better performance of FL-GUARD (our approach) compared with other baselines under each experimental setting under "IID" and "non-IID (by role)" schemes. All above observations are similar to those in previous work [1].

Table 6 demonstrates that client inactivity does pose some negative impacts on FL since both methods see a reduction in accuracy when the ratio of active clients in each round varies from 90% to 10%. Compared to FedAvg, the accuracy and local performance gain achieved by FL-GUARD is much more stable. Note the change in learning accuracy may not be monotonic to the change in the ratio of active clients. This is probably because under our test environment, where clients' data are non-IID, more active clients may slightly enhance the random disparities in reported model parameters.

Table 7 shows the negative impacts of the poisoning attacks. As the proportion of attackers increases quantitatively, the performance metrics of FedAvg reduce significantly. However, FL-GUARD is much more resilient against such negative impacts. In particular, FL-GUARD provides a high positive gain in ACC (β > 10) even when the ACC of the global model reduces by nearly 15. The results further confirm the effectiveness of FL-GUARD in adapting to local data and thereby salvaging the local performance of FL on clients.

Table 8 presents the negative impacts caused by the noises introduced by different privacy. A reduction occurs in both accuracy and local performance gain as the noises increase quantitatively from σ = 0.001 to σ = 0.005. The change in the results on SHAKE is less considerable and not strictly monotonic compared to the results on CIFAR. This indicates the better resilience of the federated LSTM against noises from differential privacy compared with the federated CNN. In contrast to FedAvg, FL-GUARD produces much better accuracy. These results show that FL-GUARD is more resilient against large noises and can make a good adaptation on client data.

### 5.4 Additional Properties of FL-GUARD

We conduct additional experiments to study the compatibility of FL-GUARD with previous techniques for tackling NFL. We also investigate the robustness of FL-GUARD against vanilla FL clients who stick to following FedAvg and do not take any NFL recovery measures. Meanwhile, we evaluate *partial layer adaptation* in FL-GUARD, namely to make local adaptation on only the top layer(s) of the model for reducing the recovery cost. All experiments in this section are tested under the default environment settings. We present the results on CIFAR here, as those on SHAKE are pretty similar.

**Compatibility with previous NFL recovery techniques.** Table 9 presents the results of FL-GUARD combining our NFL detection scheme and model adaptation with previous NFL recovery techniques, which include FedProx

**Table 9** FL-GUARD is compatible with previous techniques for tackling NFL

| Metric | ACC | $\beta$ |
|---|---|---|
| Default | 84.85 | +12.33 |
| + FedProx | 84.93 | +12.41 |
| + TrimmedMean | 84.25 | +11.73 |
| + multi-Krum | 84.69 | +12.17 |
| + K-norm | 85.59 | +12.07 |
| w/ APFL | 83.22 | +10.70 |
| w/ Ditto | 84.84 | +12.32 |

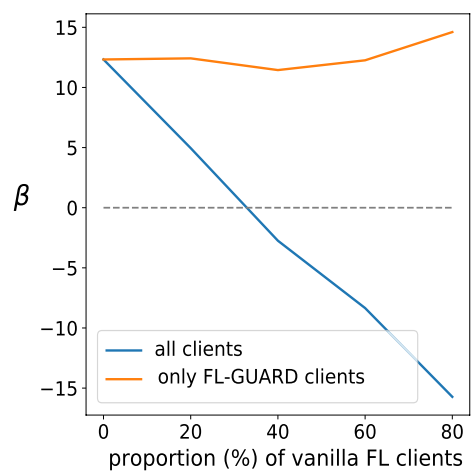

**Fig. 4** FL-GUARD is robust against vanilla FL clients

[38], TrimmedMean [31], multi-Krum [43], and K-norm [28]. We also try to replace our model adaptation approach with APFL and Ditto, which produce competitive results in Sect. 5.2. It can be seen that integrating these techniques into FL-GUARD does not significantly influence the system performance. This illustrates the flexibility of our FL-GUARD framework in adopting different kinds of NFL recovery techniques along with our NFL detection scheme (and model adaptation) for tackling NFL in a run-time paradigm.

**Robustness against vanilla FL clients.** We now investigate the robustness of FL-GUARD against vanilla FL clients who stick to following FedAvg and do not take any NFL recovery measures. We show (1) the average $\beta$ of all clients in FL and (2) the average $\beta$ of FL-GUARD clients (i.e., those following the rules in FL-GUARD to run the recovery scheme). As shown in Fig. 4, when the proportion of vanilla FL clients increases from 0% to 80%, the average $\beta$ of all clients reduces quickly below zero, indicating that the entire system's resistance to NFL is weakened by the increasing proportion of vanilla FL clients. However, the average $\beta$ of FL-GUARD clients is much more resilient against such variance, remaining high above zero even when the vanilla clients account for over 50%. These results show the robustness of FL-GUARD against vanilla FL clients and again confirm the effectiveness of our recovery scheme. Such robustness has never been reported in any previous work but is important when using a new technique to upgrade an FL system in use. The performance gains obtained by FL-GUARD clients can be persuasive evidence for attracting vanilla FL clients into our FL-GUARD community.

**Partial layer adaptation.** Finally, we show that when adopting the model adaptation for NFL recovery, the recovery costs can easily be reduced via *partial layer adaptation*, i.e., to make local adaptations to only parameters in the top-most layer(s) of the neural model on each client while freezing (namely to refrain from updating) the lower layers. Table 10 presents the results of partial layer adaptation with the different number of lower layers to be frozen ($L_{fb}$). Surprisingly, no apparent loss in accuracy is observed when the number of frozen layers increases. The results indicate that, when making model adaptation in FL-GUARD, it is safe to freeze several lower layers of neural models to reduce the adaptation costs without compromising accuracy.

**Table 10** FL-GUARD can reduce its NFL recovery costs without compromising accuracy via partial layer adaptation

| Metric | ACC | $\beta$ |
|---|---|---|
| Default | 84.85 | +12.33 |
| $L_{fb} = 1$ | 85.04 | +12.52 |
| $L_{fb} = 2$ | 84.33 | +11.81 |
| $L_{fb} = 3$ | 82.95 | +10.43 |
| $L_{fb} = 4$ | 82.06 | +9.54 |

## 6 Conclusion

This paper addressed the problem of negative federated learning (NFL). We proposed to leverage an estimated performance gain for each client from participating in FL to detect NFL at the early stage of learning. The estimated per-client performance gain was obtained efficiently by leveraging the information produced when clients train the global model. We also designed a technique for NFL recovery, which additionally learned an adapted model for each client. All these techniques are employed in a holistic framework called FL-GUARD, which tackles NFL in a run-time detection and recovery paradigm. Extensive experiments showed that FL-GUARD achieved higher accuracy than previous approaches and handled various NFL (and ideal FL) scenarios effectively and efficiently. We also showed that FL-GUARD was compatible with existing NFL recovery techniques and robust against clients who do not take any recovery measures. For future work, we are interested in exploring more effective and efficient schemes in NFL detection and recovery to enhance FL-GUARD further.

**Acknowledgements** This work was supported by the Pioneer R &D Program of Zhejiang (No. 2024C01021).